\documentclass[final]{cvpr}

\usepackage{times}
\usepackage{epsfig}
\usepackage{graphicx}
\usepackage{amsmath}
\usepackage{amssymb}

\usepackage{comment}
\usepackage{algorithmic}
\usepackage{graphicx}
\usepackage{textcomp}
\usepackage{epsfig}
\usepackage{subcaption}

\usepackage{cite}
\usepackage{enumitem}

\def\BibTeX{{\rm B\kern-.05em{\sc i\kern-.025em b}\kern-.08em
    T\kern-.1667em\lower.7ex\hbox{E}\kern-.125emX}}
    
\newcommand{\matr}[1]{\mathbf{#1}}
\DeclareMathOperator*{\minimize}{minimize}

\usepackage{multirow}
\makeatletter
\@namedef{ver@everyshi.sty}{}
\makeatother
\usepackage{tikz}
\usetikzlibrary{positioning}
\usetikzlibrary{calc}
\usetikzlibrary{fit}
\usepackage{dblfloatfix}

\usepackage{mathtools}

\usepackage[pagebackref=true,breaklinks=true,colorlinks,bookmarks=false]{hyperref}

\def\cvprPaperID{5559} % *** Enter the CVPR Paper ID here
\def\confYear{CVPR 2021}

\newtheorem{theorem}{Theorem}[section]

\newtheorem{lemma}[theorem]{Lemma}

\begin{document}

\title{Exploring the Interchangeability of CNN Embedding Spaces}

\author{David McNeely-White
\and
Ben Sattelberg
\and
Nathaniel Blanchard
\and
Ross Beveridge\\
Colorado State University\\
Fort Collins, Colorado, USA\\
{\tt\small \{david.white, ben.sattelberg, nathaniel.blanchard, ross.beveridge\}@colostate.edu}
}

\maketitle

%%%%%%%%% ABSTRACT
\begin{abstract}
  CNN feature spaces can be linearly mapped and consequently are often interchangeable. This equivalence holds across variations in architectures, training datasets, and network tasks. Specifically, we mapped between 10 image-classification CNNs and between 4 facial-recognition CNNs.
   When image embeddings generated by one CNN are transformed into embeddings corresponding to the feature space of a second CNN trained on the same task, their respective image classification or face verification performance is largely preserved. For CNNs trained to the same classes and sharing a common backend-logit (soft-max) architecture, a linear-mapping may always be calculated directly from the backend layer weights. However, the case of a closed-set analysis with perfect knowledge of classifiers is limiting. Therefore, empirical methods of estimating mappings are presented for both the closed-set image classification task and the open-set task of face recognition. The results presented expose the essentially interchangeable nature of CNNs embeddings for two important and common recognition tasks. The implications are far-reaching, suggesting an underlying commonality between representations learned by networks designed and trained for a common task. One practical implication is that face embeddings from some commonly used CNNs can be compared using these mappings.
   %We believe that this has major implications for the deep learning community, as training and retraining CNNs for the same task appears in the cases shown to produce largely interchangeable feature embeddings. 
\end{abstract}

%%%%%%%%% BODY TEXT
\section{Introduction}
\begin{figure}[t]
  \centering
  \includegraphics[width=0.45\textwidth]{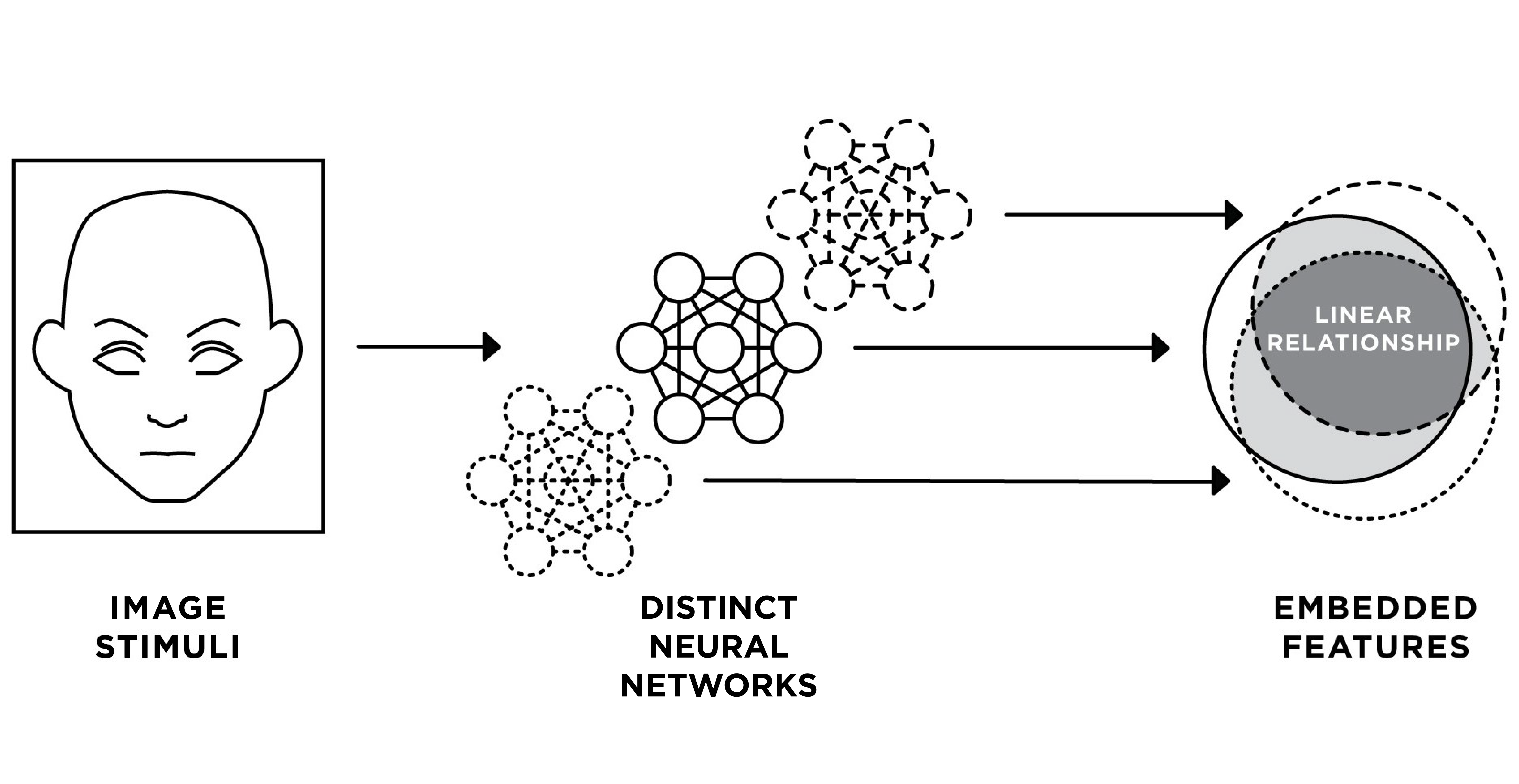}
  \caption{Distinct CNNs trained to specific tasks, for example object recognition (ImageNet) or face recognition (LFW), create embeddings which are, in many cases, interchangeable within tasks. This paper will present both analytical and empirical findings relating to the existence of linear mappings between embedding spaces. By empirically estimating these mappings we will reveal common cases where embeddings from apparently very different networks are nearly interchangeable. }  
  \label{fig:teaser}
\end{figure}
The last decade of research into neural networks might be coarsely categorized into efforts toward: 1) advancing the state-of-the-art as expressed through accuracy, and 2) better understanding and analyzing what networks learn.
Efforts toward understanding have been far ranging: from comparisons to human vision \cite{blanchard2019neurobiological,storrs2020diverse}, to investigations into the quality and content of learned features \cite{hermann2020origins}, to detailed breakdowns of what causes networks to fail \cite{dodge2016understanding}. Even research areas as distinct as style transfer~\cite{hart2020style,gatys2016image} and transfer learning \cite{oquab2014learning} are all, from a certain vantage, centered around understanding the nature of what CNNs learn and, in turn, how what they learn separates and semantically organizes information. 
Indeed, such fundamental understanding is crucial for continued improvement.
%of machine learning methods.
%At first glance, one might assume that direct comparison between embedding spaces is unlikely to be meaningful, since CNNs are a combination of millions of parameters tied to nonlinear functions. However, the significant term in the sentence above is `directly'. 

%Indeed, since these models are commonly randomly initialized and stochastically trained using perhaps millions of natural samples (e.g. in the case of ImageNet \cite{russakovsky2015imagenet}), any naive approach simply comparing embeddings be direct approach to comparing embeddings seems, and generally is, fruitless. 
Our work here focuses specifically on the embeddings generated by different CNNs. To be clear, what we will call an embedding here is the feature vector generated by the network layer just ahead of the final classifier. Often the classifier takes the form of a series of units, one per object class, and the network's final decision is expressed through a winner-take-all competition between these units. However, as we will see in the open-set face recognition task below, an alternative to a final classifier layer for a fixed set of labels is to make direct comparisons between embeddings derived from images.   

The question we take up in this work is whether their exists a linear mapping between embedding spaces generated by different CNNs trained to a common task.  What we find is that for the two cases explored in detail, object recognition and face recognition, the answer is yes. This commonality of embedding spaces is illustrated in Figure~\ref{fig:teaser}. Because we demonstrate the existence of these linear mappings, and also how estimate them, we reveal that, in a very basic sense, apparently different and incomparable embeddings from different networks are, to a great extent, interchangeable when the networks are trained to solve a common task. 

In revealing a deep commonality between embedding spaces created by very different CNNs, our work is in line with other recent and significant research trends. For example, some within the neural architecture search community have shifted their focus from improving the architecture itself to finding new data augmentation strategies \cite{cubuk2019autoaugment, zoph2019learning, cubuk2020randaugment}. There is, we believe, a growing sense that different architectures applied to the same data are, in a broad sense, converging upon very similar solutions.

The remainder of this paper is organized as follows. Related work is presented in Section~\ref{sec:related}.
Then, in Section~\ref{sec:imagenet-methods} we empirically derive mappings between 10 ImageNet-trained CNNs of varying architecture, demonstrating broad linear correspondence in a closed-set context.
Also, in Section~\ref{ssec:methods-analytic}, we identify potential concerns when using accuracy as a metric for closed-set feature space similarity.
Section~\ref{sec:methods-face} builds upon the close-set result by establishing linear correspondence in the open-set facial recognition domain. Of particular importance is that the open-set nature of face recognition means some networks we are comparing were trained on disjoint datasets and indeed disjoint sets of people. 
Finally, in Section~\ref{sec:conclusion}, we discuss both broad implications and also 
a very practical consequence for face embeddings. Namely, the fact that mappings between different embedding spaces can be estimated means identities of individuals present in multiple databases may be transferred from one to the other.

\section{Related Work}
\label{sec:related}
Many have studied neural networks for the purpose of understanding their hidden representations.
Techniques range from visualization \cite{erhan2009visualizing, zeiler2014visualizing, olah2017feature}, to semantic interpretation \cite{kim2017interpretability, zhou2016learning}, to similarity metrics. The efforts with similarity metrics are most aligned to our goals with this work, and constitute the bulk of our related work. 

Li~\etal\cite{li2015convergent} used matching algorithms to align individual units of ImageNet-trained deep CNNs to determine if networks of different initializations converge to similar representation. 
However, others have argued that drawing semantic meaning from individual bases is problematic, suggesting that semantic meaning is contained within the entire feature space, rather than individual units \cite{szegedy2013intriguing}.

Studies on representational similarity studied the full feature space using canonical correlation analysis (CCA), or variants of CCA such as SVCCA, projection weighted CCA, or centered kernel alignment. Relevant to our work, these studies have revealed a number of interesting deep neural network properties, like the tendency of layers to be learned in a bottom-up fashion \cite{raghu2017svcca}, a correspondence between networks' size and width and relative similarity \cite{morcos2018insights}, and similarity across and within networks trained on different datasets \cite{kornblith2019similarity}. The latter work is particularly relevant, as it focuses on the degree of representational similarity between networks which differ in parameterization, much as our own does. CCA methods typically look at a variety of layers and often investigate toy problems. In contrast, our work focuses solely on comparing different architectures' hidden representations using linear regression and relatively large-scale recognition tasks.

To our knowledge, only two other works have used linear regression to compare hidden deep neural network representations.
Lenc and Vedaldi \cite{Lenc2019} previously investigated linear transforms between layers of AlexNet \cite{krizhevsky2012imagenet}, VGG-16 \cite{simonyan2013deep}, and ResNet-v1-50 \cite{he2016deep} on ILSVRC2012.
However, these mappings were between spatially-sensitive convolutional layers, and thus required interpolation, which significantly factors into mapping performance.
Further, they used supervision in the form of image labels, which, in some sense, amounts to a newly-trained network.
In our work, we fit mappings using corresponding pairs of features without interpolation, and independent of semantic image labels.

McNeely-White~\etal\cite{mcneely2019inception} come closest to our methodology, providing the basis for our first set of experiments.
They found linear similarity between feature representations of ILSVRC2012-trained Inception-v4 \cite{szegedy2017inception} and ResNet-v2 152 \cite{he2016identity} models.
We map between those and 8 additional ImageNet CNNs. Further, we consider mappings between feature spaces of CNNs trained using different datasets and labels, and evaluate in an open-set context.

Recently, Roeder~\etal established a theoretical basis for the empirical linear correspondence we observe has emerged \cite{roeder2020linear}.
In the limit of infinite data, they show that a broad class of models, including deep neural networks, will converge to representations which are similar, up to a linear transformation.
They empirically demonstrate their finding using SVCCA on a variety of tasks, including a toy supervised classification problem. 
We complement this work by empirically demonstrating this effect for large CNNs of varying architecture and domains.

\section{Classifier Based Linear Maps}
\label{sec:imagenet-methods}

Our goal is to compare CNNs which vary in architecture, training dataset, or both. 
We begin with ILSVRC2012 trained networks, where all of our models are trained on the same dataset, but differ in architecture. 
Each network and its associated weights was obtained from TensorFlow Hub\footnote{\url{https://tfhub.dev/}} with the exception of Inception-v4, which was obtained from the TensorFlow-Slim GitHub repository\footnote{\url{https://github.com/tensorflow/models/tree/master/research/slim}}.
See Table \ref{tab:cropacc} for details of the networks and their respective top-1 single-crop classification accuracies on the 50,000 ILSVRC2012 validation samples as reported by Google and reproduced by us\footnote{Note: we use a crop size of 331x331, except in the case of MobileNet-v2 which is only available pretrained with a maximum crop size of 224x224, to minimize the difference in features encoded into each network's feature vectors.  This means most networks actually perform slightly better than reported by Google, although MobileNet-v2 performs marginally worse.}.

%These networks have a large variety in network architecture, 
These networks' architectures vary widely, 
from the simple residual connections of ResNets \cite{he2016deep, he2016identity}, the hand-built modules of Inception nets \cite{szegedy2015going, szegedy2016rethinking, szegedy2017inception}, the resource-constrained design of MobileNet-v2 \cite{sandler2018mobilenetv2}, to the heavily optimized NASNet and PNASNet \cite{zoph2018learning, liu2018progressive}.  Due to the significant methodological differences used for constructing these networks, it seems reasonable to assume they will partition the space used to assign class labels differently. 
%it is not unreasonable to expect that the ways in which they partition space for assigning class labels may be different.

\subsection{Classifier Based Metrics}
\label{ssec:imagenet-metrics}
\begin{table*}
\centering
    \caption{Classification accuracies of the 10 CNNs studied on the ILSVRC2012 validation set, both reported by Google and verified independently (with respective crop sizes).
    Also included for reference are the number of dimensions used in each CNN's feature space, and the total number of parameters present in the model.}
    \begin{tabular}{|l|l|l|l|l|l|l|}
    \hline
    \textbf{CNN} & Reported & Reported Crop & Ours & Our Crop & Feature Dimension & \# params \\ \hline
    Inception-v1\cite{szegedy2015going} & 69.8\% & 224x224 & 71.1\% & 331x331 & 1024 & 5M \\ \hline
    Inception-v2\cite{szegedy2016rethinking} & 73.9\% & 224x224 & 73.9\% & 331x331 & 1024 & 11M \\ \hline
    MobileNet-v2-1.4-224\cite{sandler2018mobilenetv2} & 74.9\% & 224x224 & 74.6\% & 224x224 & 1792 & 7M \\ \hline
    ResNet-v1-152\cite{he2016deep} & 76.8\% & 224x224 & 78.8\% & 331x331 & 2048 & 60M \\ \hline
    ResNet-v2-152\cite{he2016identity} & 77.8\% & 224x224 & 78.7\% & 331x331 & 2048 & 60M \\ \hline
    Inception-v3\cite{szegedy2016rethinking} & 78.0\% & 299x299 & 78.9\% & 331x331 & 2048 & 24M \\ \hline
    Inception-v4\cite{szegedy2017inception} & 80.1\% & 299x299 & 80.4\% & 331x331 & 1536 & 43M \\ \hline
    Inception-ResNet-v2\cite{szegedy2017inception} & 80.4\% & 299x299 & 81.2\% & 331x331 & 1536 & 56M \\ \hline
    NASNet-Large\cite{zoph2018learning} & 82.7\% & 331x331 & 82.7\% & 331x331 & 4032 & 89M \\ \hline
    PNASNet-Large\cite{liu2018progressive} & 82.9\% & 331x331 & 82.9\% & 331x331 & 4320 & 86M \\ \hline
    \end{tabular}
    \label{tab:cropacc}
\end{table*}

% \begin{figure}
% \centering
% \includegraphics[width=0.45\textwidth]{figures/method.png}
% \caption{Illustration of how two linear-classifier-based CNNs can be tested for linear equivalence}
% \label{fig:method}
% \end{figure}

To better lay out our approach, let us develop a format that clearly establishes our intended meaning by outlining features/embeddings and their connection to object labeling. 

Each network presented in Table~\ref{tab:cropacc} may  be considered a function, $v:\mathbb{R}^{w \times h \times 3} \to \mathbb{R}^{1000}$. The domain consists of $w \times h$ pixel RGB images and the range is the $\mathbb{R}^{1000}$ label space; each dimension represents a network activation corresponding to one of the $1,000$ ILSVRC2012 object labels.  The function $v$ can be further divided into two parts:  
\begin{equation}
    v(x) = \matr{C} f(x).
    \label{eqn:classifier-network-definition}
\end{equation}
Here, $f:\mathbb{R}^{w \times h \times 3} \to \mathbb{R}^d$ is a highly nonlinear function mapping from the input image space to a $d$-dimensional feature/embedding space. The matrix $\matr{C} \in \mathbb{R}^{1000 \times d}$ is a linear classifier that transforms a feature vector into class label activations. The argmax of these activations is typically used to determine the predicted class.  Note that many traditional neural network systems include some form of bias that would appear to complicate this definition, but by mapping to projective space (always appending a $1$ to vector $f$), the last column of $\matr{C}$ can be used to represent the bias.

%We are interested in calculating to what extent a linear mapping converts between the features of two networks.  

Given two networks, $v_A$ and $v_B$ defined by
\begin{align}
\begin{split}
    v_A(x) &= \matr{C}_A f_A(x) \\
    v_B(x) &= \matr{C}_B f_B(x),
\end{split}
\end{align}
with features of system $v_A$, $f_A \in \mathbb{R}^{d_A}$, and features of system $v_B$, $f_B \in \mathbb{R}^{d_B}$, we are interested in the extent to which the functions $f_A$ and $f_B$ behave similarly.  These functions are unlikely to have identical output or be directly comparable, but we can investigate the extent to which they have a linear relationship, i.e. does there exist a matrix $\matr{M}_{A\to B} \in \mathbb{R}^{d_B \times d_A}$ such that
\begin{equation}
    f_B(\cdot) \approx \matr{M}_{A\to B} f_A(\cdot).
    \label{eqn:map-target}
\end{equation}

Naively, calculating distance between the features $f_B(x)$ and $f_A(x)$ for each sample in the ILSVRC2012 dataset using something like Euclidean distance appears tempting, but for a variety of reasons it is not helpful. Most obvious is the problem of possible representational permutations. It is well understood that two networks of identical architecture may permute feature dimensions, yet be in all other ways be equivalent. Additionally, even when setting aside the permutation problem it is also known that distances in high dimensional spaces can sometimes obfuscate results~\cite{bellman1957markovian}. Further, comparing feature spaces is complicated by possible differences in the number of dimensions from one network to the next. 

We do want to mention that a promising avenue to work around such problems is to employ methods such as canonical correlation analysis~\cite{hotelling1992relations,morcos2018insights}. However, because such methods are invariant to linear transforms, they give insight into similarity while not serving our goal of constructing and understanding the $\matr{M}_{A\to B}$ mappings.

Due to these potential issues with common distance metrics, a useful proxy for us to measure similarity is the top-1 accuracy of network combinations: what accuracy is attained by the system using the linear mapping:
\begin{equation}
    v_{A\to B}(x) = \matr{C}_B \matr{M}_{A \to B} f_A(x).
    \label{eqn:mapped-classifier-network}
\end{equation}

We will show in Section~\ref{ssec:methods-analytic} some conditions such that when $\matr{C}_A$ and $\matr{C}_B$ are known, a matrix $\matr{M}_{A \to B}$ may be calculated directly from them. However, we will choose first to explore the broader case where such knowledge is not available and hence $\matr{M}_{A \to B}$ must be estimated empirically. 
%See Figure~\ref{fig:method} for an illustration.

\subsection{Empirically Calculated Mapping}
\label{ssec:methods-empirical}

Although Euclidean distance is potentially a poor measure of similarity between spaces as a whole, it may be used as optimization metric to efficiently construct empirical $\matr{M}_{A \to B}$ mappings.  Our calculation of the mapping is a ridge regression problem over all the 1.3 million images in the training set, $X$, using the Euclidean norm $||\cdot||_2$ and the Frobenius norm $||\cdot||_F$:
\begin{equation}
\label{eqn:ridge}
    \minimize_{\tilde{\matr{M}}_{A\to B}} \sum_{x_i \in X} ||\tilde{\matr{M}}_{A\to B} f_A(x_i) - f_B(x_i)||_2
    \, + \,
    ||\tilde{\matr{M}}_{A\to B}||_F
\end{equation}
The resulting matrix $\tilde{\matr{M}}_{A\to B}$ minimizes total point-wise Euclidean distance between the feature spaces.

We then use $\tilde{\matr{M}}_{A\to B}$ for pairs of networks to construct the mapped networks $v_{A\to B}$ as defined by Equation~\ref{eqn:mapped-classifier-network}. The recognition accuracy for the mapped networks is then calculated over the ILSVRC2012 test set.  This process is completed for each pairwise combination of the 10 pre-trained CNNs and the results are summarized in Table~\ref{tab:cropacc}.  One thing to emphasize here is that neither the ground truth image labels nor the $\matr{C}_A$ and $\matr{C}_B$ matrices are used in when estimating the linear mappings $\tilde{\matr{M}}_{A\to B}$.

One additional control system is also considered. The control is simply the PNASNet-Large network using only the initial randomly generated weights. We add this control to provide some backdrop against which to empirically assess the relative difference between the mapped networks designed and trained to achieve high recognition accuracy. %The control will also be a useful point of reference in Section~\ref{ssec:methods-analytic}.
%we further explore that fact that just because a linear mapping may be estimated it is not always the case that the mapping ensures comparably good results; mapping features from a poor network cannot magically induce excellent recognition performance. 

%110 mappings (omitting the 11 identity mappings).  We note that that the neither the ground truth image labels nor the $\matr{C}_A$ and $\matr{C}_B$ are used in this training process.  Accuracy results for these mapped systems are shown in Table~\ref{tab:mapped-classifer-eval}.

\subsection{Classifier Based Mapping Results}
\label{ssec:imagenet-results}
\begin{table*}
    \caption{
    Classification accuracies of 121 inter-CNN linear maps.
    %Each cell represents a single, independent linear mapping between the feature extractor from the \textbf{row CNN} and the classifier from the \textbf{column CNN}.
    Each cell represents a single instance of Equation~\ref{eqn:mapped-classifier-network} with system $v_A$ corresponding to the \textbf{row CNN} and system $v_B$ corresponding to the \textbf{column CNN}.
    %The number in large font in each cell indicates the classification accuracy of this hybrid CNN on the 50k ILSVRC2012 validation set.
    The number in large font in each cell indicates the accuracy of this hybrid CNN.
    %Note: this means the diagonal for the first 10 CNNs shows the performance of an identity mapping, equivalent to the performance of original CNNs in Table~\ref{tab:cropacc}.
    Diagonal elements correspond to identity mappings, so they are the original network accuracies from Table~\ref{tab:cropacc}.
    The number in small font indicates the percent change from the original unmapped row/source CNN (i.e. the value in that row which belongs to the diagonal).
    The darker the shade of red, the greater the performance penalty introduced by the mapping, relative to the feature extractor's own classifier. 
    The last row and column, gray background, presents comparisons with a random untrained control CNN that is the PNASNet Large architecture using randomly initialized weights. 
    }
    \includegraphics[width=\textwidth]{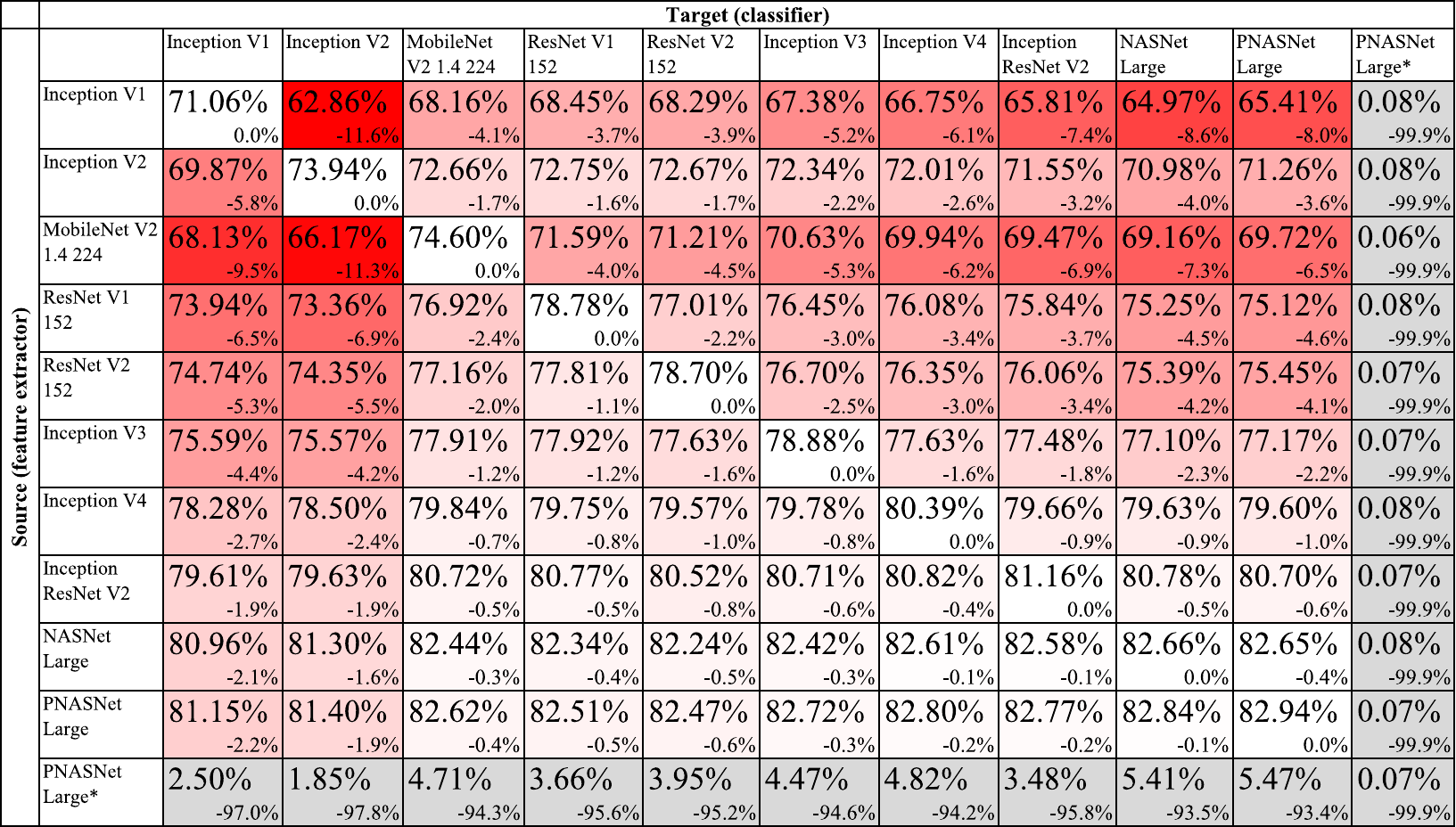}
    \label{tab:mapped-classifer-eval}
\end{table*}
 
Table~\ref{tab:mapped-classifer-eval} presents the performance of all combinations of CNN features and back-end classifiers. 
%Row headings indicate the model whose feature extractor is used as the feature vector \textbf{source}, while column headings indicate the model whose classifier is used as \textbf{destination} after mapping the features of the source to the destination.
%Each cell contains the classification accuracy achieved by the independently trained mapping between feature extractor and classifier.
%Values along the diagonal are identity mappings, and indicate the accuracy of the unmodified source CNN as in Figure~\ref{tab:cropacc}.
%The small number indicates the percent change in accuracy from the unmodified \textbf{source} CNN (i.e. the value in that row which belongs to the diagonal).
%Cells are shaded according to this percent change, so that the darker the shade, the greater performance penalty introduced by the mapping.
All accuracies are reported using the ILSVRC2012 validation set, which was unseen by all CNNs and mappings during training.
In addition a randomly-initialized PNASNet-Large, denoted by *, is also included.

% While the reduction in classification accuracy introduced by each mapping is never zero, the greatest is only an 11.6\% reduction.
% Comparing across feature extractors (row-wise), mappings from the Inception-v1 and MobileNet-v2-1.4-224 feature vectors incur a consistently larger penalty (though no greater than 11.6\%).
% This may be because these two models are the smallest, consisting 5 million and 7 million parameters, respectively.
% Still, every mapping incurs relatively little penalty when used to transform between different feature spaces.
% These mappings are effective enough to retain the vast majority of features' expressiveness, even when used for classification.

% Comparing across classifiers (column-wise), mapped feature vectors which are classified by Inception-v1's and Inception-v2's classifiers incur the largest and most consistent penalties.
% Again there may be many explanations, though these two models uniquely use the smallest feature space at 1024 dimensions.
When comparing column-wise (feeding different features into the same classifier), accuracies are sometimes \textit{increased} when compared to the classifier CNN's own features.
That is, even though the mappings were trained to reproduce the classifier CNN's less expressive features, some additional helpful information is passed through the mapping.
As an example, see ResNet-v2-152's column and unmapped accuracy of 78.70\%.
This unmapped accuracy is taking ResNet-v2's features and passing them into ResNet-v2's classifier.
When instead fed a linear transformation of PNASNet's features, ResNet-v2's classifier produces 82.46\% accuracy.
This is an increase in the ResNet-v2 classifier's accuracy of 3.76\%!
Of course, when analyzing the effects of mappings on the \textit{source} CNN's performance (i.e. the fashion that cells were shaded in Table~\ref{tab:mapped-classifer-eval}), {\bf no feature vectors actually become more discriminative} when mapped.
Additionally, the sharpest dropoffs in accuracy occur for the lowest-accuracy networks.  This suggests that as architectures become complex enough to solve the ILSVCR2012 dataset well, they are converging to similar solutions as predicted by Roeder~\etal~\cite{roeder2020linear}.  

The central question is whether the features learned by each trained CNN studied are equivalent.
In every case, the percent change in classification accuracy introduced by using another CNN's features is no worse than -11.6\% (and in most cases, much better).
Indeed, the median percent change over all mappings is only -1.90\%.
This strongly suggests that the features learned by one CNN are also learned by every other, though some have learned somewhat superior variations.
%Indeed, some CNNs seem to have learned to express features nearly \textit{identically}, such as NASNet and PNASNet which are penalized by only 0.01\% when using the former's features in the latter's classifier.

One additional item of note is that the PNASNet-Large with randomly initialized weights improves from the baseline random accuracy when mapped to the features of other networks. To be clear, the improvement is dramatic from one perspective, jumping by as much as two orders of magnitude. However, it is lackluster at best from another perspective, reaching only about $5$ percent accuracy in the best case. This finding lends credence to the view that even just the CNN architecture itself, independent of learned weights, encodes useful information.  Indeed, this property of networks has been  explored in more depth in a number of different contexts including fast architecture search --- see~\cite{saxe2011random, pinto2009high, jarrett2009best, giryes2016deep}.

\subsection{Analytic Mapping}
\label{ssec:methods-analytic}

An important caveat to this accuracy-based comparison is that there is a vacuous sense in which the accuracy of the mapped systems can be high while similarity of the feature spaces may be low.
\begin{lemma}
Consider a mapping $\matr{C}_A \in \mathbb{R}^{d \times d_A}$ and a mapping $\matr{C}_B \in \mathbb{R}^{d \times d_B}$ that is tall (i.e. $d < d_b$) and of full row rank (i.e. rank $d$).  Then, there exists a matrix $\matr{M}_{A\to B}$ such that
\begin{equation}
    \matr{C}_A = \matr{C}_B \matr{M}_{A\to B}
\end{equation}
where $\matr{M}_{A\to B}$ can be defined as
\begin{equation}
    \matr{M}_{A\to B} = \matr{C}_B^+ \matr{C}_A.
\end{equation}
Here $\matr{C}_B^+$ is the Moore-Penrose inverse of $\matr{C}_B$.
\end{lemma}
This isn't necessarily an astonishing fact --- it is simply a statement that any matrix with full row rank has a right inverse and as such we can write
\begin{equation}
    \matr{C}_B \matr{M}_{A\to B} = \matr{C}_B \matr{C}_B^+ \matr{C}_A = I\matr{C}_A = \matr{C}_A
\end{equation}
However, it has the following unfortunate consequence:
\begin{theorem}
\label{thm:vacuous-accuracy}
Given two systems,
\begin{align}
\begin{split}
    v_A(x) &= \matr{C}_A f_A(x) \\
    v_B(x) &= \matr{C}_B f_B(x)
\end{split}
\end{align}
as defined by Equation~\ref{eqn:classifier-network-definition}, if $\matr{C}_B$ is of full rank then then there exists a matrix $\matr{M}^*_{A\to B}$ such that for all $x \in \mathbb{R}^{w \times h \times 3}$,
\begin{equation}
    \matr{C}_A f_A(x) = \matr{C}_B \matr{M}^*_{A\to B} f_A(x).
\end{equation}
where $\matr{M}_{A\to B}^*$ can be defined as
\begin{equation}
    \matr{M}_{A\to B}^* = \matr{C}_B^+ \matr{C}_A.
\end{equation}
Here $\matr{C}_B^+$ is the Moore-Penrose inverse of $\matr{C}_B$.
\end{theorem}
All ILSVRC2012 networks considered here have classifiers that are of full rank. 

A few consequences of this theorem:
\begin{itemize}[noitemsep, topsep=0pt]
    \item For any two linear-classifier based neural networks, there exists a linear mapping such that $v_{A\to B}$ has accuracy exactly equal to that of $v_{A}$
    \item This linear mapping and accuracy is independent of $f_B$ and the accuracy is independent of $\matr{C}_B$.
\end{itemize}
This means that although high linear similarity in feature space implies high accuracy after mapping, we cannot simply say that high accuracy after a linear mapping implies high linear similarity.

As an example, consider a system which when given an image generates a random feature vector, and which uses a fixed full rank classifier.  This system has random performance on ILSVRC2012, and features from a real neural network cannot be mapped to its changing ``features'' in any meaningful sense.  However, when ``mapping'' to its features using the above analytic mapping, we achieve the performance of the source network.  

This does not necessarily mean this analytic mapping does not transform between feature spaces well when given features from two networks --- it just means that it may not (and clearly does not in certain situations like the random classifier) and that the metric we are using does not always tell us when the metric may succeed or fail.

To avoid this issue in our empirical mappings, we perform our computation of the matrix $\tilde{\matr{M}}_{A\to B}$ without knowledge of the classifiers and entirely between the two feature spaces.  Additionally, we are optimizing for Euclidean distance between feature spaces, rather than the resulting mapped network accuracy.  This means that such a mapping is taking advantage of similarity in feature space when being calculated, rather than accuracy.  

Additionally, we have included the random control system described above to highlight the distinction being drawn here. In particular, the poor performance as seen in Table~\ref{tab:mapped-classifer-eval} when mapping to this control system demonstrates that the optimization problem we are solving to map between feature spaces likely does not fall prey to the potential issues in the analytic solution shown to exist by Theorem~\ref{thm:vacuous-accuracy}. 

Perhaps the simplest way to summarize what we've seen so far, is to recognize the following asymmetry of implications. To start, when Theorem~\ref{thm:vacuous-accuracy} is applicable there {\bf always exists} an exact linear mapping between two CNNs. However, the existence of a linear mapping is not always a sufficient condition for concluding features are similar. The introduction of the random control addresses this latter point. This distinction also helps spur our interest in studying linear mappings between CNNs in contexts where there is no common classifier, and as such Theorem~\ref{thm:vacuous-accuracy} does not apply. 

\section{Open-set Linear Maps}
\label{sec:methods-face}

%In Section~\ref{sec:imagenet-methods}, we described how to establish linear correspondence between two classifier-based CNNs and test that correspondence by obtaining accuracy using mapped features with the linear classifier.

Here we explore how to estimate linear mappings between feature spaces for CNNs constructed to solve open-set problems. Face recognition has been chosen for these experiments because it is both a compelling application and also because it is a task which, by its very nature, demands generalization to new labels when moving from training to testing.  So while a linear classifier may be used as the final layer during training, 
the classifier is discarded after training. During operation and validation face verification is rooted in measuring similarity between pairs of face embeddings derived from novel unseen images of people not seen during training~\cite{arcface, facenet}. This is in our opinion an intriguing domain in which to seek out and find linear mappings between feature spaces associated with CNNs of differing architecture trained on different data. 

Four common face-recognition networks are selected for study here. 
  Two share the same architecture and are trained on different datasets using center loss by David Sandberg and are available on GitHub\footnote{\url{https://github.com/davidsandberg/facenet}.} \cite{facenetGithub}.  The other two have different architectures but were trained on the same dataset by Kuan-Yu Huang and are available on GitHub\footnote{
\url{https://github.com/peteryuX/arcface-tf2}.}\cite{arcfaceGithub}.  All four use multi-task CNNs (MTCNN) to align and crop face images before feeding the result into the network and all networks normalize their feature outputs to the unit hypersphere.  Network details are available in Table~\ref{tab:face-model-desc} and the training datasets are described in Table~\ref{tab:face-dataset-desc}.  Additional training details such as learning rate schedules, optimizers, and specific software details are documented at the source repositories. 
Importantly, the four networks differ from one another by either architecture, training dataset, or both.

These pre-trained models are used to generate pairs of feature embeddings over all 6,000 image pairs specified in the LFW dataset \cite{labeledfacesinthewild}.
Accuracy is measured by performing 10-fold cross-validation on aligned image embeddings as specified by the \textit{Image Restricted, labeled outside data} protocol documented in LFW \cite{labeledfacesinthewild}.  All four models we consider were internally validated to within 0.01\% of stated accuracies on LFW.

\begin{table}[t]
  \centering
  \renewcommand{\arraystretch}{1.1}
  \begin{tabular}{l|l|l}
  Dataset & \# individuals & \# images \\\hline
  VGGFace2 & 9131 & 3.31 M \\
  CASIA-WebFace & 10,575 & 494,414 \\
  MS-Celeb-1M & 100,000 & 10 M \\\hline
  LFW & 5,749 & 13,233
  \end{tabular}
  \caption{The datasets used in our experiments. The first three datasets were used to train networks used in our experiments. The fourth, LFW, was used to test mappings between feature spaces.}
  \label{tab:face-dataset-desc}
\end{table}

\begin{table*}[!b]
  \centering
  \small
  \renewcommand{\arraystretch}{1.5}
  \begin{tabular}{l|l|l|l|l}
  Name & \textbf{CNN} & \textbf{Loss function} & \textbf{Training Dataset} & \textbf{Accuracy on LFW}  \\ \hline
  \textbf{Model-IC}\cite{facenetGithub} & InceptionResNetV1 \cite{szegedy2017inception} & Softmax + Center \cite{centerLoss} & CASIA-WebFace \cite{CASIAwebface} & 99.03\% $\pm$ 0.42\%  \\
  \textbf{Model-IV}\cite{facenetGithub} & InceptionResNetV1 \cite{szegedy2017inception} & Softmax + Center \cite{centerLoss} & VGGFace2 \cite{vggface2} & 99.47\% $\pm$ 0.37\%  \\
  \textbf{Model-MM}\cite{arcfaceGithub} & MobileNetV2 \cite{sandler2018mobilenetv2} & ArcFace \cite{arcface} & MS-Celeb-1M \cite{ms-celeb-1m} & 98.70\% $\pm$ 0.50\%  \\
  \textbf{Model-RM}\cite{arcfaceGithub} & ResNet-v2-50 \cite{he2016identity} & ArcFace \cite{arcface} & MS-Celeb-1M \cite{ms-celeb-1m} & 99.40\% $\pm$ 0.46\% 
  \end{tabular}
  \caption{Configuration and accuracy of each model. A shortened name is provided for use in other tables. Note these accuracy values are calculated by internal verification and differ from the stated values for each model's source publication and also that these models are not the official versions associated with the original publications. 
  }
  \label{tab:face-model-desc}
\end{table*}
\begin{table*}[!b]
  \centering
  \small
  \renewcommand{\arraystretch}{1.5}
  \begin{tabular}{l|l|l|l|l}
  Name & \textbf{CNN} & \textbf{Loss function} & \textbf{Training Dataset} & \textbf{Accuracy on LFW}  \\ \hline
  \textbf{Model-IC} & InceptionResNetV1 & Softmax + Center & CASIA-WebFace & 99.03\% $\pm$ 0.42\%  \\
  \textbf{Model-IV} & InceptionResNetV1 & Softmax + Center & VGGFace2 & 99.47\% $\pm$ 0.37\%  \\
  \textbf{Model-MM} & MobileNetV2 & ArcFace & MS-Celeb-1M & 98.70\% $\pm$ 0.50\%  \\
  \textbf{Model-RM} & ResNet-v2-50  & ArcFace & MS-Celeb-1M & 99.40\% $\pm$ 0.46\% 
  \end{tabular}
  \caption{Configuration and accuracy of each model. A shortened name is provided for use in other tables. Note these accuracy values are calculated by internal verification and differ from the stated values for each model's source publication and also that these models are not the official versions associated with the original publications. 
  }
  \label{tab:face-model-desc}
\end{table*}

\subsection{Open-Set Based Metrics}
\label{ssec:face-metrics}

\begin{figure*}[h!]
  \centering
  \begin{tikzpicture}[scale=0.8, transform shape, node distance = 0.75cm and 0.75cm]
  \node (image-1) {\includegraphics[width=2cm]{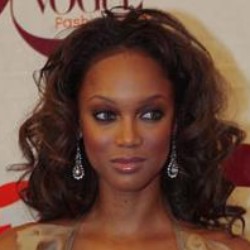}};
  \node[below = 0.3cm of image-1] (image-2) {\includegraphics[width=2cm]{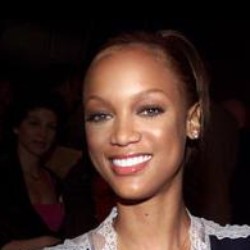}};
  
  \node[above = 0.3cm of image-1,align=center] (image-label) {Image/Video\\pair};
  
  \node[right = 1cm of image-1] (net-1) {$f_A \begin{pmatrix} & \includegraphics[width=2cm]{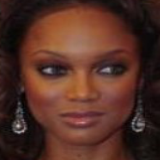} & \end{pmatrix}$};
  \node[right = 1cm of image-2] (net-2) {$f_B \begin{pmatrix} & \includegraphics[width=2cm]{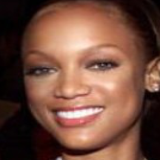} & \end{pmatrix}$};
  
  \node[align=center] (net-label) at (image-label -| net-1) {Network\\outputs};
  
   \draw [->] (image-1) -- (net-1) node[midway,above] {\footnotesize{MTCNN}};
   \draw [->] (image-2) -- (net-2) node[midway,above] {\footnotesize{MTCNN}};
   
   \node[right = of net-1] (embed-1) {$\begin{bmatrix} 0.051 \\ 0.009 \\ \vdots \\ -0.040 \\ -0.027 \end{bmatrix}$};
   \node[right = of net-2] (embed-2) {$\begin{bmatrix} 0.048 \\ -0.001 \\ \vdots \\ -0.042 \\ 0.062 \end{bmatrix}$};
   
   \node[align=center] (embed-label) at (image-label -| embed-1) {Feature\\embeddings};
   
   \draw [white] (net-1) -- (embed-1) node[pos=0.5,fill=white,text=black] {=};
   \draw [white] (net-2) -- (embed-2) node[pos=0.5,fill=white,text=black] {=};
   
   \node[right = of embed-1] (map-1) {$\mathbf{M_{A\to B}}^T \begin{bmatrix} 0.051 \\ 0.009 \\ \vdots \\ -0.040 \\ -0.027\end{bmatrix}$};
   \draw [->] (embed-1) -- (map-1);
   
   \node[right = of map-1] (mapped-embed-1) {$\begin{bmatrix} 0.062 \\ -0.005 \\ \vdots \\ -0.019 \\ -0.011\end{bmatrix}$};
   
   \draw [white] (map-1) -- (mapped-embed-1) node[pos=0.5,fill=white,text=black] {=};
   
   \node[align=center] (map-label) at (image-label -| mapped-embed-1) {Mapped\\feature};
   
   \node (distance) at (17.5, -1.5) {$d\left(\begin{bmatrix} 0.062 \\ -0.005 \\ \vdots \\ -0.019 \\ -0.011\end{bmatrix}, \begin{bmatrix} 0.048 \\ -0.001 \\ \vdots \\ -0.042 \\ 0.062 \end{bmatrix}\right) \leq \tau$};
   % TODO: how to make this automatic and actually centered? :(
   
   \node[align=center] (distance-label) at (image-label -| distance) {Distance for\\thresholding};
   
   \draw [->] (mapped-embed-1) -- (distance);
   \draw [->] (embed-2) -- (distance);
  \end{tikzpicture}
  \caption{The pipeline used for evaluating the linear mappings between face recognition CNNs. Both networks use separate multi-task CNNs (MTCNN) to detect and align the faces. The networks are then used to generate feature embeddings in $\mathbb{R}^{512}$. The linear mapping is used to map the feature embedding of the source network ($f_1$) into the feature space of the target network ($f_2$). Finally, the distance $d$ between the embeddings is computed and compared against the matching threshold $\tau$; often distance is simply one minus the cosine of the angle between embedding vectors. 
  }
  \label{fig:pipeline}
\end{figure*}
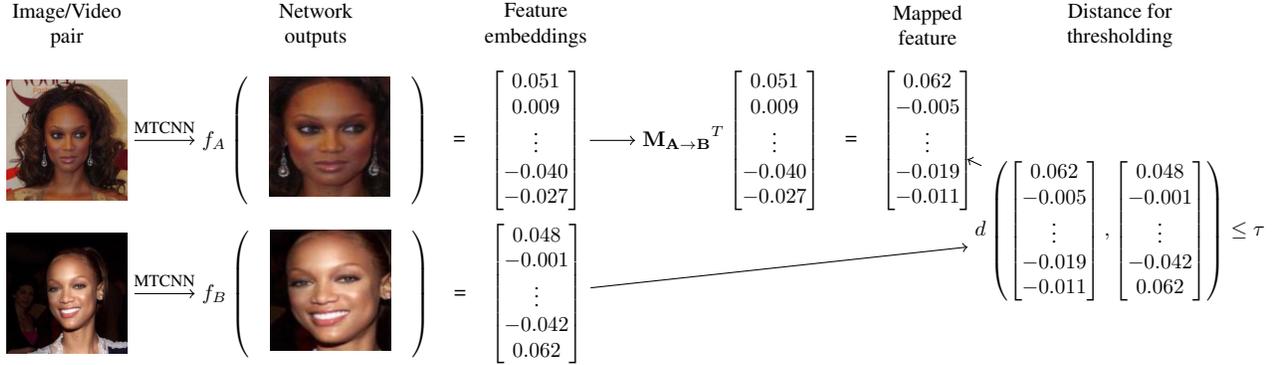 %fig:pipeline

In contrast to the classifier based systems defined in Equation~\ref{eqn:classifier-network-definition}, these open-set networks can be written as $v:\mathbb{R}^{w \times h \times 3} \to \mathbb{R}^{d}$, that for a given $w \times h$ pixel RGB image $x \in \mathbb{R}^{w \times h \times 3}$, can be defined as:
\begin{equation}
    \label{eqn:open-set-network-definition}
    v(x) = f(x)
\end{equation}
for the associated feature extractor $f: \mathbb{R}^{w \times h \times 3} \to \mathbb{R}^d$  with the condition that $||f(\cdot)||_2|| = 1$.  Typically the unit length of the feature is not enforced explicitly by the network, so the final step of feature extraction is normalizing it to the unit hypersphere.  For the four networks studied here, the dimensionality of the feature space, $d$, is always 512.  Two faces $x_1$ and $x_2$ are considered to be of the same person if their cosine distance,
\begin{equation}
\label{eqn:distance}
d(x_1, x_2) = 1 - f(x_1) \cdot f(x_2),
\end{equation}
is smaller than some established threshold $\tau$.

For testing in the 10-fold validation setting of LFW, for each fold a cosine distance threshold $\tau$ is found which maximizes accuracy (i.e. labels the most image pairs correctly as matched or mismatched) over the other nine folds.  Using the remaining validation partition, accuracy is measured using the same threshold.   This process is repeated for each of 10 folds given by the LFW dataset in \texttt{pairs.txt} and averaged to produce a final accuracy measurement.

As before, we are interested in calculating the extent to which a linear map converts between the features of two networks.   For two systems
\begin{equation}
\begin{split}
    v_a(x) &= f_a(x) \\
    v_b(x) &= f_b(x)
\end{split}
\end{equation}
as defined in Equation~\ref{eqn:open-set-network-definition}, does there exist a matrix $\matr{M}_{A\to B} \in \mathbb{R}^{512 \times 512}$ such that 
\begin{equation}
    f_B(x_2) \approx \matr{M}_{A\to B} f_A(x_1)
\end{equation}
is satisfied for input images $x_1, x_2 \in \mathbb{R}^{w \times h \times 3}$ corresponding to the same individual.  This approach maps clusters corresponding to individuals in one network to the same clusters in another network, rather than mapping points in feature space specifically. Note that we do not explicitly normalize the result of $\matr{M}_{A\to B} f_A(x)$.

Given such a matrix, we can calculate the cosine distances between corresponding image pairs, as
\begin{equation}
d_{A\to B}(x_1, x_2) = 1 - \frac{(\tilde{M}_{A \to B}f_A(x_1)) \cdot f_B(x_2)}{||\tilde{M}_{A \to B} f_A(x_1)||_2 }.
\end{equation}
The denominator here is due to the fact that we do not normalize $\matr{M}_{A\to B} f_A(x)$.  The distance $d_{A\to B}$ is used to calculate an optimal threshold, $\tau$, which is then used with $d_{A\to B}$ to determine if unseen pairs of images are of the same individual as in standard LFW validation. This process is demonstrated in Figure~\ref{fig:pipeline}. 
Once a mapping is calculated and threshold $\tau$ determined, we use classification accuracy as a measure of linear similarity between feature extractors $f_A$ and $f_B$.

%This method does not fall prey to the vacuous sense in which classification accuracy can be satisfied after a linear mapping when using classifer-matrix based networks.  We are actually looking at similarity in feature space, so the projection to the shared classification space is not relevant as it was in Theorem~\ref{thm:vacuous-accuracy}.  In this case accuracy does fully imply similarity.

\subsection{Empirically Calculated Mapping}
\label{ssec:face-methods}

We fit mappings in a fashion similar to Subsection~\ref{ssec:methods-empirical}.
Instead of fitting linear maps between pairs of embeddings corresponding to the same \textit{image}, however, we fit mappings during the LFW evaluation process using pairs of embeddings corresponding to the same \textit{individual}.
To be more precise, during the evaluation of one fold of LFW, the set of $m$ matched image pairs that have both images representing the same individual, $(x_i\in\mathcal{X}, y_i\in\mathcal{Y}), i=1,...,m$, from the nine training folds are used as training data for fitting the linear mapping.  $\mathcal{X}$ and $\mathcal{Y}$ do not overlap in specific images.
For reference, each LFW fold contains 300 matched and 300 mismatched pairs, so $m=9*300=2,\!700$.

We calculate mappings by solving the ridge regression problem over these image pairs similarly to Equation~\ref{eqn:ridge}:
\begin{equation}
    \minimize_{\tilde{\matr{M}}_{A\to B}} \sum_{i=1}^m ||\tilde{\matr{M}}_{A\to B} f_A(x_i) - f_B(y_i)||_2
    \, + \,
    ||\tilde{\matr{M}}_{A\to B}||_F
\end{equation}
This produces a matrix $\tilde{\matr{M}}_{A\to B}$ which minimizes the distance between clusters corresponding to individuals in feature space.

We then use these estimated mappings $\tilde{\matr{M}}_{A\to B}$ to classify image pairs in the remaining validation fold as illustrated in Figure~\ref{fig:pipeline}.
Validation accuracy scores are averaged over 10 folds to produce a single accuracy value.
This process is completed for each permutation of the 4 models in Table~\ref{tab:face-model-desc} to produce 12 mappings (omitting 4 identity mappings).
Face verification accuracies for these mapped systems are shown in Table~\ref{tab:lfw-mapping-accuracy}.

\begin{table}[]
  \centering
  \renewcommand{\arraystretch}{1.5}
  
  \caption{The open-set accuracy of each face recognition CNN when mapped, using \textbf{LFW}. The format of this table is identical to Table~\ref{tab:mapped-classifer-eval}, including the same color scale. 
  %(darkest = 11.6\% change). 
%   Diagonal elements correspond to the unmodified accuracy of each model (see Table~\ref{tab:face-model-desc}). Off-diagonal elements correspond to the accuracy obtained when comparing features across networks, with the ``Source'' model's features mapped by linear transformation to approximate the ``Target'' model's features. The maximum drop in accuracy introduced by any mapping is 1.0\%. The number in small font indicates the percent change from the unmapped row/source CNN. Cells are shaded according to this percent change in accuracy, using the same color scale as Table~\ref{tab:mapped-classifer-eval}. Note that for the LFW validation set, 50\% accuracy corresponds to random chance. 
  }
  
  \includegraphics[width=0.45\textwidth]{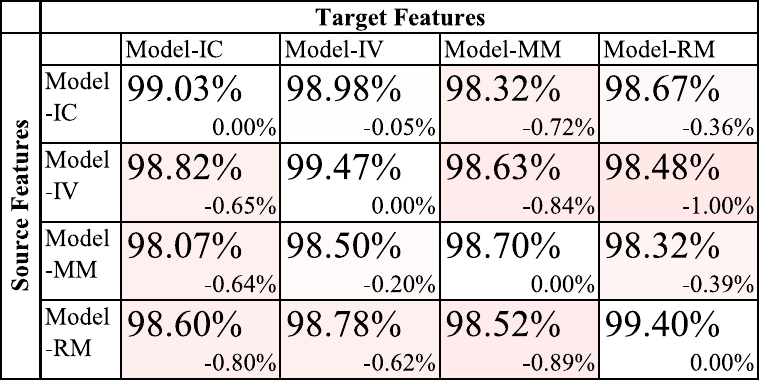}
  \label{tab:lfw-mapping-accuracy}
\end{table}

\subsection{Open Set Mapping Results}
\label{sec:face-results}

The loss in accuracy introduced by linear mapping is at most 1.0\% on LFW. 
In our experiments, Model-MM is outperformed by other models by a large margin, which may account for the poor performance of mappings to and from its feature space. %Indeed, excluding Model-MM reduces the maximum mapping performance penalty on YTF by more than half. Further, the images in YTF are generally lower quality due to motion blur or distance, which may account for greater disparities in performance. 
Regardless, the ability for one network's features to robustly replicate another's demonstrates a near-linear equivalence of the feature spaces. 
Although some information does not transfer, the vast majority of information that is encoded by each network is equivalent. 
This finding is in contrast to previous results that methods such as feature fusion increase the accuracy of the networks~\cite{bansal2018deep, bodla2017deep}. However, we note that the small amount of information not encoded by this linear transform may result in the increases in accuracy revealed in those experiments. 

One additional check is direct interchangeability between these feature spaces without mapping.  As discussed in~\cite{li2015convergent, szegedy2013intriguing}, direct interchangeability between feature spaces is unlikely to exist.  
We confirm that by using an identity mapping as $\matr{M}_{A\to B}$.  When we do so, we achieve near random accuracy. This confirms that our results demonstrate that empirically estimated linear mappings are able to map reliably between embedding spaces, and consequently suggest one way in which the spaces may be considered largely interchangeable. 

\section{Conclusion}
\label{sec:conclusion}
The work of Roeder~\etal~\cite{roeder2020linear} mentioned above is highly significant, suggesting a theoretical basis for concluding that neural networks applied to a common task will, in the limit, converge upon a common representation up to a linear transformation. 
%Roeder~\etal's empirical analysis addressed supervised classification tasks, self-supervised pretraining, and multi-task natural language generation pretraining. Their supervised learning tasks focused on a small-scale problem not on the scale of the ILSVRC2012 or face recognition datasets presented here.  Consequently, 
We believe our work nicely complements that of Roeder~\etal by demonstrating the existence of linear mappings in two significant object-recognition domains and well-recognized CNNs.  We believe future work exploring the presence, or perhaps absence in some cases, of linear mappings using new ML models and more varied task domains will be both interesting and rewarding.   

Our work also has some significant practical implications.  Let us highlight one relating to the use of face embeddings to encode identity; as part of a large multi-institution federally funded research program, (Name will be added here after double blind reviewing is over), the face-embedding mapping technique introduced in Section~\ref{sec:methods-face} was used to reveal the concealed identity of a person of interest in one database using a named instance in an entirely separate database. At least in this experiment, our ability to establish a mapping demonstrates the risk in believing that withheld identities cannot be discovered using cross database linkages. This is just one example of what can be done once it is understood that: 1) these mappings seem to be common, and 2) they can be uncovered empirically.  

%\section{Discussion}

%Funding acknowledgement!! (DARPA AIDA)

%Acknowledgement that this work was adapted in part from David's thesis
\paragraph*{Acknowledgements}
%This work was adapted in part from David McNeely-White's Master's Thesis \cite{mcneely2020same}.
This work was supported by the US Defense Advanced Research Projects Agency (DARPA) and the Army Research Office (ARO) under contracts \#W911NF-15-1-0459 and \#FA8750-18-2-0016.
Thank you to Michelle Wern for designing Figure~\ref{fig:teaser}.

{\small
\bibliographystyle{ieee_fullname}
\bibliography{MAIN}
}

\end{document}